\documentclass[10pt,twocolumn,letterpaper]{article}
\usepackage[accsupp]{axessibility}
\usepackage[algorithms]{wacv}      

\definecolor{wacvblue}{rgb}{0.21,0.49,0.74}
\usepackage[pagebackref,breaklinks,colorlinks,allcolors=wacvblue]{hyperref}

\def\wacvPaperID{*****} 
\def\confName{WACV}
\def\confYear{2026}

\title{Efficient Visual Question Answering Pipeline for Autonomous Driving via Scene Region Compression}

\author{Yuliang Cai\\
University of Southern California\\
{\tt\small caiyulia@usc.edu}
\and
Dongqiangzi Ye\\
XPeng\\
{\tt\small eowinye@gmail.com}
\and
Zitian Chen\\
XPeng\\
{\tt\small tankche2@gmail.com}
\and
Chongruo Wu\\
XPeng\\
{\tt\small chongruo@gmail.com}}

\begin{document}
\maketitle
\begin{abstract}
Autonomous driving increasingly relies on Visual Question Answering (VQA) to enable vehicles to understand complex surroundings by analyzing visual inputs and textual queries. Currently, a paramount concern for VQA in this domain is the stringent requirement for fast latency and real-time processing, as delays directly impact real-world safety in this safety-critical application. However, current state-of-the-art VQA models, particularly large vision-language models (VLMs), often prioritize performance over computational efficiency. These models typically process dense patch tokens for every frame, leading to prohibitive computational costs (FLOPs) and significant inference latency, especially with long video sequences. This focus limits their practical deployment in real-time autonomous driving scenarios. To tackle this issue, we propose an efficient VLM framework for autonomous driving VQA tasks, \textbf{SRC-Pipeline}. It learns to compress early frame tokens into a small number of high-level tokens while retaining full patch tokens for recent frames. Experiments on autonomous driving video question answering tasks show that our approach \textbf{achieves 66\% FLOPs} reduction while maintaining comparable performance, enabling VLMs to operate more effectively in real-time, safety-critical autonomous driving settings.
\end{abstract}

\section{Introduction}
Autonomous driving increasingly depends on large vision–language models (VLMs) to enhance multimodal understanding by bridging visual and textual modalities through large-scale pretraining. Early models \cite{clip,BLIP-2,LLaVA} have demonstrated that combining powerful vision encoders with large language models enables strong performance on a variety of multimodal tasks, such as image captioning, image-text retrieval, visual question answering (VQA), etc. Extending these ideas, video question answering has emerged as a promising paradigm for domains where temporal reasoning and causal understanding are critical. In particular, autonomous driving is an application field where answering natural language queries about dynamic traffic scenes can evaluate and enhance the reasoning capabilities of the perception system. Recent benchmarks \cite{driveLM,nusceneQA,lingoQA} exemplify this trend, providing large-scale datasets that frame autonomous driving challenges—such as detecting pedestrians, recognizing traffic signals, or predicting vehicle intent—as visual-based question answering problems.

\begin{figure}[t]
\centering
\includegraphics[width=\columnwidth]{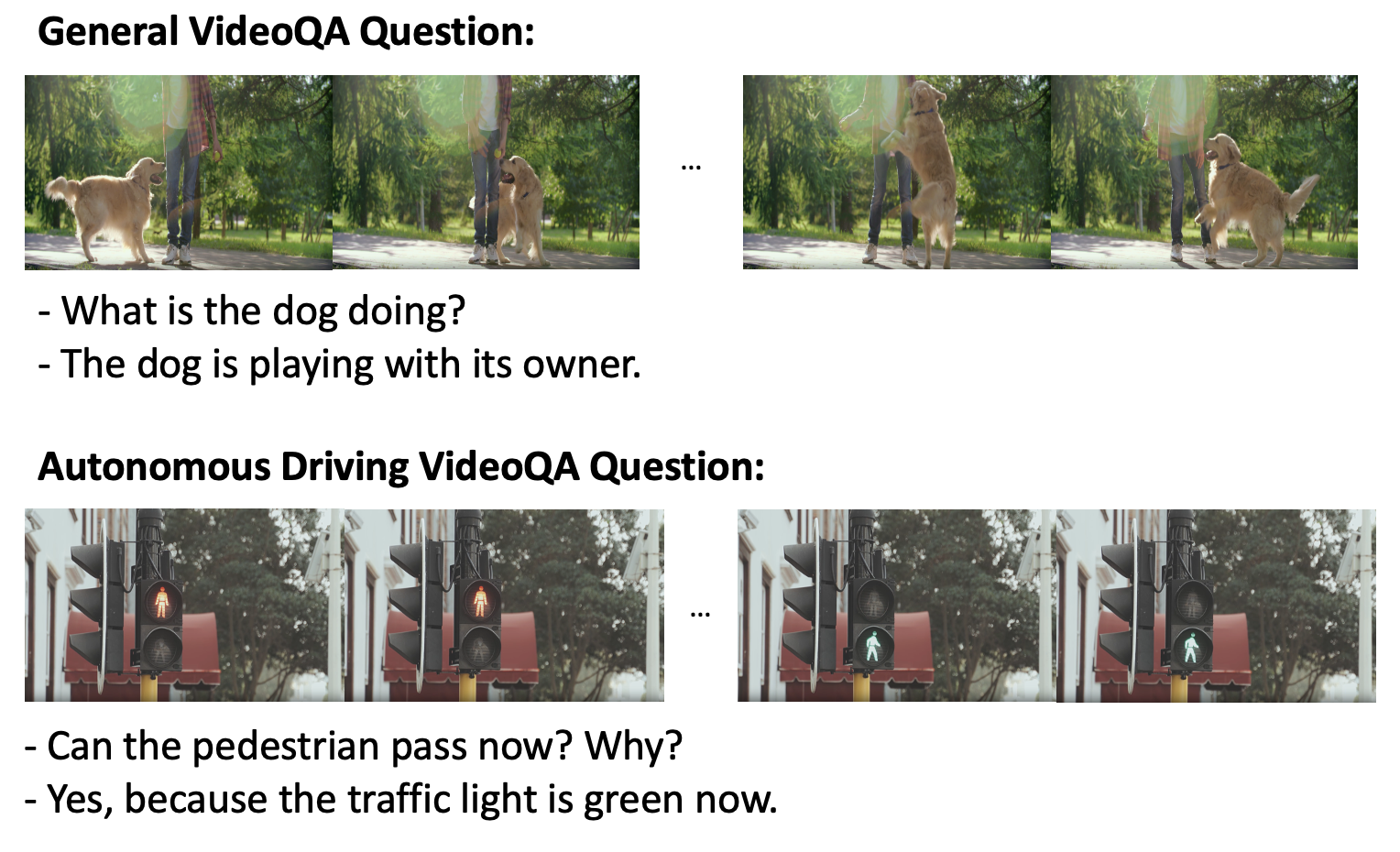}
\caption{Difference between general-purpose video question answering and autonomous driving video question answering. While the question of the general-purpose video question answering covers the information from all the frames, the question of the autonomous driving video question answering mostly focuses on the later frames due to the timeliness of the autonomous driving video.}
\label{stage1train}
\end{figure}

Despite these advances, existing VLMs face serious efficiency challenges when applied to autonomous driving. Modern models often contain billions of parameters and rely on dense patch tokenization of each video frame, which can result in hundreds of visual tokens per frame \cite{internvl2.5,qwen2vl}. While such designs are suitable for offline benchmarks, they are not suitable to safety-critical driving scenarios where real-time perception and reasoning are essential. Autonomous driving systems must react within strict latency budgets—often a few hundred milliseconds—when making decisions such as braking for pedestrians or changing lanes in dense traffic. In such contexts, the high floating-point operation (FLOP) requirements and large memory footprint of current VLMs present a major barrier to deployment. Even modest delays in processing visual streams can degrade safety, underscoring the importance of designing VLMs that are both accurate and computationally efficient.

To address this challenge, recent works have begun to explore efficient multimodal modeling strategies. Approaches such as token pruning, adaptive visual tokenization, and dynamic resolution scaling \cite{vila,qwen2vl} attempt to reduce the number of tokens passed to the language decoder, thereby lowering the dominant FLOPs cost. However, these methods only focus on the token-efficient design for general video question answering, without considering the unique property of the \textbf{autonomous driving} video question answering (VideoQA). In general VideoQA task~\cite{videoqa}, the questions are highly diverse—ranging from summarization and description to captioning and decision making—and may require detailed information from arbitrary temporal segments of the video. For example, the questions could be \textit{"what is the cat doing?"}, \textit{"Why is the kid crying?"}, and \textit{"Is the dog barking at the first ten seconds?"} In contrast, the questions from autonomous driving VideoQA \cite{lingoQA,malla2023drama} are relatively homogeneous, focusing on the real-time perception (eg. \textit{what is the traffic light color now?}) and the decision making (eg. \textit{what should be the next-step action of the vehicle?"}), which highly depend on the visual information from the later frames in the video. 

Based on the uniqueness of the autonomous driving VideoQA, we propose an efficient VLM framework for video question answering of autonomous driving, \textit{SRC-Pipeline}, which leverages frame-level token compression. We hypothesis that \textbf{in the domain of real-time traffic scene analysis and decision making, later frames are typically more informative than earlier frames.} By summarizing early frames into compact high-level tokens while retaining dense patch tokens for later, contextually critical frames, our method reduces the computational burden of VLM inference with only slightly drop of reasoning accuracy. This design significantly lowers FLOPs in the language decoder—the primary computational bottleneck—while maintaining the ability to reason about both long-term temporal context and fine-grained spatial detail. We demonstrate that our approach achieves competitive accuracy with substantially reduced computation on driving-focused VideoQA benchmarks, highlighting its potential for real-time deployment in latency-sensitive autonomous driving systems.

\section{Related Works}
Visual Question Answering (VQA) has emerged as a crucial area of research for enhancing the interpretability, trustworthiness, and overall capability of autonomous driving (AD) systems. By enabling vehicles to answer natural language questions about their surrounding environment, VQA bridges the gap between raw sensor data and human-understandable reasoning, a critical step for safety-critical applications like autonomous driving \citep{lingoQA,rekanar2024optimizingvisualquestionanswering,atakishiyev2023explainingautonomousdrivingactions}. Recent advancements in Vision-Language Models (VLMs) and Multimodal Large Language Models (MLLMs) have significantly propelled VQA research in AD. 

These models, by synthesizing linguistic and visual data, promise a new era of intelligent and efficient transportation systems capable of deep environmental understanding and natural language interaction. LMDrive \citep{lmdrive} proposes an LLM-based end-to-end driving framework on close-loop scenario by integrating multi-modal sensor data with natural language instructions, enabling interaction with humans and navigation software in realistic instructional settin DriveGPT4 \citep{drivegpt4}, on the other hand, develops an multi-modal large language model to offer interpretation of the vehicle's action and predict the low-level control signal in an end-to-end manner. 
TOKEN \citep{tian2024tokenizeworldobjectlevelknowledge} proposes a novel Multi-Modal Large Language Model that tokenizes the world into object-level knowledge, which aims to address long-tail events in autonomous driving by producing condensed and semantically enriched representations of the scene. 

The methods mentioned above primarily focuses on VLM performance in autonomous driving, yet often overlooks efficiency, which is crucial for real-time, safety-critical operations and constrained onboard computational resources. To tackle the efficiency issue, VTS \citep{vts} focuses on using CNN-based model to adaptively identify the key frame and reduce the less informative tokens in the autonomous driving video. FastDriveVLA \citep{fastdriveVLA} proposes a plug-and-play module for VLMs to perform token pruning. Instead of using visual token similarity or video-text attention, it designs a reconstruction-based pruning strategy using MAE-style pixel reconstruction technique.

However, while the token pruning methods can effectively reduce the redundant visual token and save bandwidth for LLM computation, they also introduce extra computation to identify the less informative and redundant tokens, which makes such approaches less efficient. As a result, we propose a new temporal-based token compression strategy that can significantly reduce the amount of visual token based on the timeliness of the frame, instead of introducing extra computation, to save the FLOPs.

\begin{figure*}[]
\centering
\includegraphics[scale=0.5]{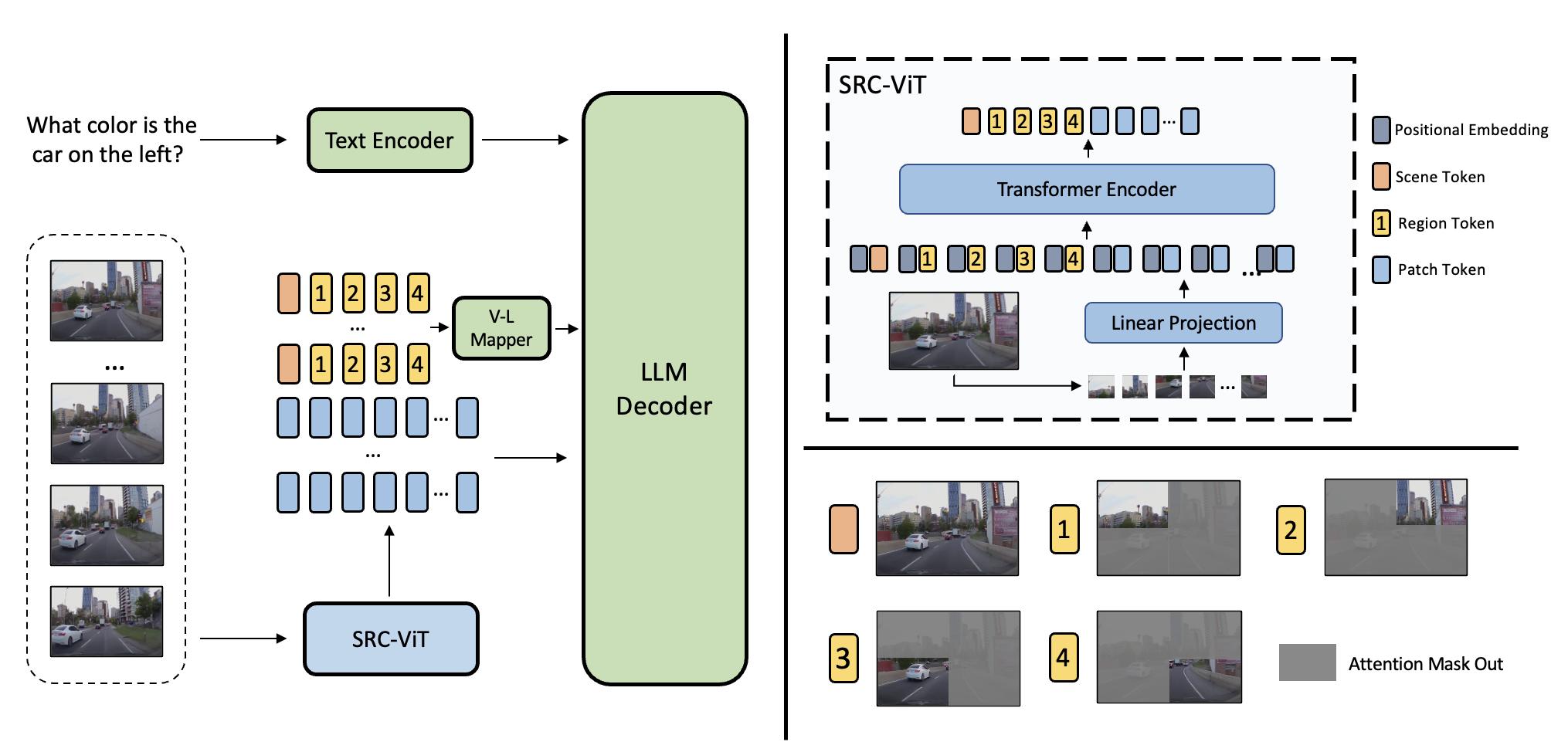}
\caption{The architecture of SRC-Pipeline. \textit{(left):} The overall architecture of SRC-Pipeline. Video frames are sent to the SRC-ViT to produce both the scene-region token and the patch token, where the early frames are represented by the scene-region tokens and the later frames are represented by the patch tokens. \textit{(upper right):} The architecture of the SRC-ViT. Learnable scene and region tokens are attached to the path tokens from linear projection, and the positional embedding is applied to both the scene region tokens and the patch tokens. \textit{(lower left):} The region attention mask. For each of the region tokens, the attention of it to the patch tokens that do not belong to the specific region will be masked out.}
\label{fig:main_Diagram}
\end{figure*}

\section{Problem Description}
Our study focuses on \textbf{video question answering (VideoQA) for autonomous driving}, where a model must reason over driving videos and preserving detailed visual information in a computational efficient manner. Given a dataset $\mathcal{D}=\{(V_i,q_i,a_i)\}_{i=1}^{N}$, where each sample consists of a driving clip $V_i=(I_1,I_2,...,I_T)$ with $T$ frames, a question 
$q_i$, and a corresponding ground truth answer $a_i$. A text encoder, $f_{text}$ maps the question, $q_i$, into a semantic feature representation:
\begin{equation}
    Q_i=f_{text}(q_i).
\end{equation}
Meanwhile, a visual encoder $f_{vis}$ extracts multi-level visual features from each video frame $I_t$:
\begin{equation}
    V_t=f_{vis}(I_t)=[C_t,P_t].
\end{equation}
Specifically, it produces two sets of tokens:
\begin{itemize}
    \item a compact contextual representation $C_t\in \mathbb{R}^{m\times d}$, where $m$ is the size of contextual tokens, which summarizes high-level semantics of the frame, and
    \item a patch token representation $P_t\in \mathbb{R}^{n\times d}$, where $n$ is the token length for the frame, capturing detailed spatial information.
\end{itemize}
To reduce computational overhead while retaining fidelity, we adopt a hybrid token selection strategy: we represent the first $m$ frames with the contextual representation. For the earlier frames ($t \leq m$), we substitute the patch tokens $P_t$ with their counterpart $C_t$, while for later frames we preserve the patch tokens. This design ensures that the model focuses its bandwidth on the more informative parts of the video, which are often concentrated in later temporal segments, while keep the global context information of the whole video, like weather, traffic condition, etc.
The final multimodal input to the language decoder $f_{dec}$ is thus constructed as 
\begin{equation}
    X=(Q,C_1,C_2,...,C_m,P_{m+1},...,P_T).
\end{equation}
The decoder $f_{dec}$ integrates visual and textual representations to generate the predicted answer $\hat{a}$.

\section{Methodology}
We introduce our proposed \textbf{s}cene-\textbf{r}egion \textbf{c}ompression pipeline (\textit{SRC-Pipeline}) for efficient autonomous driving video question answering, the pipeline is shown in left side of Figure~\ref{fig:main_Diagram}. \textbf{\textit{SRC-Pipeline} is a visual token compression technique based on the large VLMs which aims to efficiently solve the VideoQA task for autonomous driving.} The key novelty of \textit{SRC-Pipeline} is the \textbf{s}cene-\textbf{r}egion \textbf{c}ompression \textbf{vi}sion \textbf{t}ransformer (\textit{SRC-ViT}), which is the substitution of the VLM's original visual encoder that outputs not only the patch tokens of the video frame, but the scene-region high-level compression tokens for the entire frame and four regions of the frame. We integrate the fine-tuned \textit{SRC-ViT} into the vision language model to form the \textit{SRC-pipeline}. In this section, we first introduce the architecture of our \textit{SRC-ViT} in Sec \ref{sec:srcViT}, the \textit{SRC-pipeline} in Sec \ref{sec:srcPipeline}, we will then discuss our two-stage training strategy and objectives in Sec \ref{4.2.1} and Sec~\ref{4.2.2}.

\begin{figure}[t]
\centering
\includegraphics[width=\columnwidth]{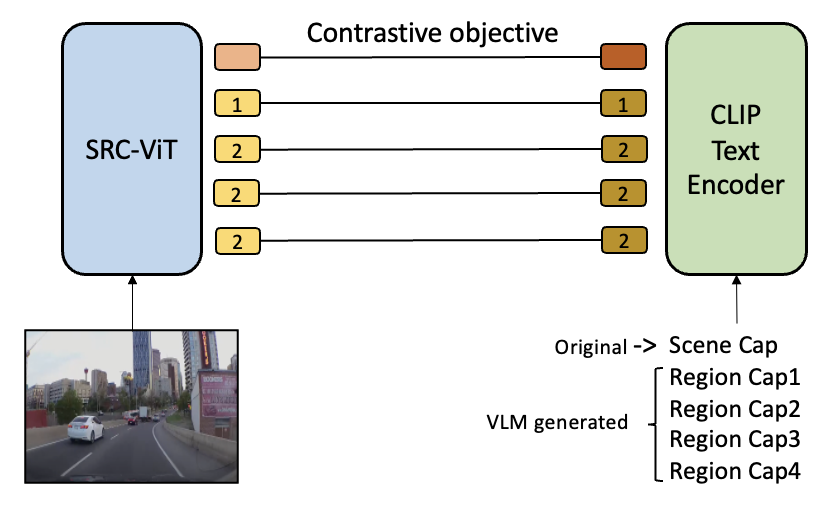}
\caption{The stage one training pipeline. The SRC-ViT is paired with CLIP's text encoder. For every image, one scene caption and four region captions are encoded via the text encoder, and the [CLS] token of each is aligned with the image's scene token and region tokens for contrastive learning.}
\label{stage1train}
\end{figure}

\subsection{Scene Region Compression for Vision Transformer}
\label{sec:srcViT}
We introduce the Scene Region Vision Transformer (\textit{SRC-ViT}), shown in the upper right of Figure~\ref{fig:main_Diagram}, a novel visual encoder designed to improve the computational efficiency of video question answering (VideoQA) for autonomous driving. The core idea of \textit{SRC-ViT} is to replace less informative patch tokens from earlier frames with compact scene-region compression tokens and reduce FLOPs while preserving essential spatial–temporal information.

Unlike the the visual encoders of the general-purpose VLMs
\cite{wang2024qwen2vlenhancingvisionlanguagemodels,internvl2.5} which input hundreds of dense patch tokens per frame into the language decoder, \textit{SRC-ViT} generates both fine-grained patch representations and high-level compression tokens. Inspired by the success of the \textit{[CLS]} token in CLIP \citep{clip}, our design extends this idea by introducing a multi-scale set of learnable tokens that capture global and localized information at different levels. Specifically, each frame is augmented with one scene token and four region tokens, corresponding to the upper-left, upper-right, lower-left, and lower-right quadrants. This enables the model to encode both holistic semantics and localized spatial details in a compressed form.

Given an input frame $I$, the patch embedding layer $E_p$ produces a sequence of patch tokens
\[
P=\{P_1,P_2,...,P_n\}=E_p(I).
\]
We then augment these tokens with a set of learnable high-level tokens, including a scene token $S$ and four region tokens ($R_{ul}$,$R_{ur}$,$R_{ll}$,$R_{lr}$). The combined input to \textit{SRC-ViT}'s transformer encoder $E_t$ is
\[
Z=[S,R_{ul},R_{ur},R_{ll},R_{lr},P_1,P_2,...,P_n]+ T^{pos},
\]
where $T^{pos}$ denotes positional embeddings.

To ensure that each region token only summarize information from its designated spatial partition, we modify the self-attention mask $M\in \{0,1\}^{(n+5)\times(n+5)}$, shown in the lower right of Figure~\ref{fig:main_Diagram}. Specifically, for a region token R, attention is restricted to the subset of patch tokens $P_r$ corresponding to its spatial quadrant:
\[
Attn(R,P_j)=0, \text{if} \ P_j\notin P_r.
\]
This constraint enforces spatial grounding, ensuring that the learned region tokens encode localized compression features align with specific areas of the frames. The transformer encoder then outputs a set of encoded features:
\[
O=E_t(Z)=[\hat{S},\hat{R_{ul}},\hat{R_{ur}},\hat{R_{ll}},\hat{R_{lr}},\hat{P_1},\hat{P_2},...,\hat{P_n}],
\]
where each element represents the updated embedding of its corresponding input token. The resulting compression tokens ($\hat{S},\hat{R_{ul}},\hat{R_{ur}},\hat{R_{ll}},\hat{R_{lr}}$) provide a compact and semantically rich summary of the frame, allowing the VLM pipline to substitue dense early-frame patches with these tokens while retaining detailed patch tokens for later, more informative frames.

\subsection{Token Compression VLM Pipeline}
\label{sec:srcPipeline}

Building on the proposed SRC-ViT, we design the \textit{SRC-Pipeline} to efficiently integrate compressed visual tokens into the vision–language model (VLM) for autonomous driving video question answering. The central motivation is that in real-time driving scenarios, later frames often contain more critical details for perception and decision making, whereas earlier frames primarily provide contextual semantics. To exploit this property, we substitute dense patch tokens from early frames with compact scene–region compression tokens, thereby reducing the FLOPs cost while preserving the semantics. 

Consider a driving video composed of $N$ frames, {$V_1,V_2,...,V_n$}. For each frame $V_i$, the \textit{SRC-ViT} produces two types of representations:
\begin{itemize}
    \item \textbf{Compression token} $C_i$, consisting of one scene token and four region tokens.
    \item \textbf{Patch tokens} $P_i$, capturing the fine-grained spatial details.
\end{itemize}
The full visual feature sequence is thus
\[
\mathcal{V}=[(C_1,P_1),(C_2,P_2),...,(C_T,P_T)].
\]
To reduce computational overhead, we represent the first $M$ frames using only the compression tokens {$C_1,C_2,...,C_M$}, , while preserving the full patch tokens {$P_{M+1},P_{M+2},...,P_T$} for the later, more informative frames. Since the high-level compression tokens must be compatible with the language decoder’s feature space, we introduce a vision–language mapper $M_{vl}$, which projects each $C_i$ into a shared multimodal embedding space:
\begin{equation}
    C'=M_{vl}(C_i).
\end{equation}
The resulting visual input to the language decoder is
\[
\mathcal{V}'=[C'_1,C'_2,...,C'_M,P_{M+1},P_{M+2},...,P_N].
\]
Finally, the visual features $\mathcal{V}'$ are fused with the encoded text representation of the question,
\begin{equation}
    T=E_q(q),
\end{equation}
where $E_q$ is the text encoder. The multimodal input to the decoder is
\begin{equation}
    I_d=[\mathcal{V}',T],
\end{equation}
which is processed by the LLM decoder $E_d$ to generate the final answer \textbf{A}.

To effectively incorporate the proposed scene–region compression strategy into the VLM, we design a two-stage training pipeline. This strategy first fine-tunes the \textit{SRC-ViT} to have meaningful high-level compression tokens and then adapts the language model to interpret these tokens in a multimodal reasoning context.

\subsubsection{Stage One: SRC-ViT Fine-tuning}
\label{4.2.1}

Since the original ViT backbone lacks high-level scene and region tokens, we train these tokens from scratch using contrastive objectives. Specifically, each image is paired with one scene caption and four region captions to provide text-level supervision at multiple spatial scales, which is shown in Figure~\ref{stage1train}. The captions are generated by a large vision–language model \citep{zhang2025videollama3frontiermultimodal} to ensure semantic consistency.

Given an image $I$, we associate it with a set of captions:
\[
\mathcal{S}(I)=\{c_s,c_{ul},c_{ur},c_{ll},c_{lr}\},
\]
where $c_s$ is the scene-level caption and $c_{ul},c_{ur},c_{ll},c_{lr}$ are the region-level captions. The SRC-ViT produces a scene token $S$ and four region tokens {$R_{ul},R_{ur},R_{ll},R_{lr}$}. Each token is aligned with the [CLS] embedding of its corresponding caption from the CLIP text encoder via a contrastive loss:

\begin{equation}
    \mathcal{L}_{total} = \frac1{2}\mathcal{L}_{scene} + \frac{1}{2}\mathcal{L}_{region},
\end{equation}
where the $\mathcal{L}_{scene}$ is the contrastive loss between the image and the scene-level caption, and the $\mathcal{L}_{region}$ is the contrastive loss between the image and the four caption-level captions. To distance the semantic features of similar scenarios and make the region-wise image-text matching more accurate, we compute $\mathcal{L}_{region}$ by contrasting one region image features with all the other region text features in the batch, and vice versa,  including the regions in the same frame.

\subsubsection{Stage Two: VLM Pipeline Fine-tuning}
\label{4.2.2}
After fine-tuning SRC-ViT, we freeze its parameters and proceed to adapt the vision–language mapper and the LLM decoder. The mapper $M_{vl}$ projects the learned high-level tokens into a multimodal embedding space that is compatible with the decoder. During this stage, the decoder is fine-tuned jointly with $M_{vl}$ so that it can interpret both dense patch tokens and compressed scene–region tokens in downstream VideoQA tasks. This process enables the LLM to seamlessly integrate compact early-frame features with detailed later-frame features, preserving reasoning quality while improving efficiency.

\section{Experiments}
\begin{table*}[h]
\centering
\small
\setlength{\tabcolsep}{6pt}
\renewcommand{\arraystretch}{1.2}
\begin{tabular}{c||c|c|c}
\toprule
\textbf{Model} & \textbf{Frames}  & \textbf{Ling-Judge}  & \textbf{BLEU} \\
\midrule
LingoQA & 5 & 60.8 & 15.00 \\
SRC-Pipeline & 5 & 59.10 & 12.55 \\
Qwen2VL & 5 & 58.20 & 13.46   \\
\midrule
SRC-Pipeline & 2 & 58.20 & 12.25 \\
SRC-Pipeline & 1 & 57.28 & 12.09 \\
LingoQA & 1 & 57.00 & 14.21  \\
LLaVA & 1 & 59.00 & 12.50\\
BLIP-2 & 1 & 52.2 & 13.0 \\
\bottomrule
\end{tabular}
\caption{Comparison experiment results on the LingoQA dataset.}
\label{compare}
\end{table*}

\begin{table*}[h]
\centering
\small
\setlength{\tabcolsep}{6pt}
\renewcommand{\arraystretch}{1.2}
\begin{tabular}{c||c|c|c|c}
\toprule
\textbf{Model} & \textbf{Frames}  & \textbf{Ling-Judge}  & \textbf{BLEU} & \textbf{FLOPs (\%)} \\
\midrule
SRC-Pipeline & 5 & 59.10 & 12.55 & 100\% \\
SRC-Pipeline & 1 & 57.28 & 12.09 & 33.3\%   \\
SRC-Pipeline (s only) & 1 & 56.42 & 11.18 & 33.3\%\\
SRC-Pipeline (avg. pooling) & 1 & 56.20 & 10.99 & 33.3\% \\
SRC-Pipeline (no sr) & 1 & 56.00 & 11.02 & 33.3\% \\
\bottomrule
\end{tabular}
\caption{Ablation study result on SRC-Pipeline components.}
\label{ablation}
\end{table*}
To demonstrate the performance of SRC-Pipeline on autonomous driving video question answering tasks, we evaluate SRC-Pipeline on the LingoQA \citep{lingoQA} dataset, and present the ablation study to examine the effectiveness of each designed component. 
\subsection{Experiment Setup}
To evaluate the performance of SRC-Pipeline, we choose the LingoQA as the evaluation benchmark \citep{lingoQA}, which contains 419.9k autonomous driving-related QA pairs. It evaluates the model's performance on four aspects, including scene description, action prediction, justification and attention. Besides the common N-Gram-based metrics such as BLEU \cite{BLEU}, LingoQA introduces a semantic-based metric, $Lingo$-$Judge$, such that it adopt fine-tuned Bert-based model as classifier to evaluate the semantic similarity between the generated answer and the ground truth answer. 

We choose the QWen2VL-7B \citep{qwen2vl} as the VLM backbone of SRC-Pipeline, which couples a 675M‑parameter Vision Transformer with a 7.6B‑parameter language decoder. To encode the scene token and the region tokens via ViT, we apply the same positional embedding to them as to the patch tokens. Notably, the QWen2VL model applies Multimodal Rotary Position Embedding (M‑RoPE) as the positional embedding of its LLM decoder input, unifying positional encoding across time, x index and y index in visual domain. To fit the high-level tokens into the M-RoPE, we modify the M-RoPE of the scene and region tokens' $x$ and $y$ position embedding to be the center of its focused area. For example, for the region token of a frame which focuses on $m^{th}$ to $n^{th}$ patch token for row and $i^{th}$ to $j^{th}$ patch token for column, the $x$ and $y$ index of it will be:
\begin{equation}
    p_x = \frac{(m+n)}{2},
    p_y = \frac{(i+j)}{2}.
\end{equation}
The temporal index of the scene region tokens are the same of the patch tokens in the same frame.

For the baseline methods, we adopt several SOTA VLMs, including LingoQA \citep{lingoQA}, LLaVA\citep{LLaVA}, BLIP-2\citep{BLIP-2}, QWen2VL\citep{qwen2vl}, and fine-tune them on LingoQA dataset.

\subsection{Experiment Result}
We present our experiment results on LingoQA in table\ref{compare}, where the column $\textbf{Frames}$ represents the number of frames preserved in the video. While LingoQA only contains 5 frames per video, $\textbf{Frames}$ of 5 indicates all the video frames are used as visual input, while $\textbf{Frames}$ of 1 indicates that only the latest one frame is used as the visual input. Across LingoQA video QA with varying input frames, our SRC-Pipeline consistently demonstrates strong performance relative to the QWen2VL backbone it builds upon. In the 5-frame setting, SRC-Pipeline surpasses QWen2VL on both Ling-Judge (59.10 vs. 58.20) and BLEU (12.55 vs. 13.46, noting that our method narrows the BLEU gap while clearly improving the judgment-based metric), indicating that adding our pipeline yields tangible gains over the base model when leveraging full video context. Although the best overall 5-frame results come from LingoQA (Ling-Judge 60.8, BLEU 15.00), its 1-frame performance (Ling-Judge 57.00, BLEU 14.21) falls below our 1-frame SRC-Pipeline variants, showing that LingoQA’s advantage diminishes when temporal context is constrained. 

\subsection{Ablation Study}
We analyze the effect of scene- and region-level compression in the SRC-Pipeline, with results shown in Table 2. Using all five frames without compression achieves the highest performance (Ling-Judge 59.10, BLEU 12.55) but at the cost of full computation (100\% FLOPs). When applying compression to the first four frames and keeping only the last frame in full detail, the FLOPs are reduced to 33.3\% while maintaining strong accuracy (Ling-Judge 57.28, BLEU 12.09). To isolate the contributions of scene and region tokens, we evaluate two variants: one using only the scene token for compressed frames (s only) and one removing scene–region compression entirely (no sr). Both variants result in performance drops (Ling-Judge 56.42/56.00, BLEU 11.18/11.02), demonstrating that region tokens provide important localized detail that complements the global semantics from scene tokens. To demonstrate that the high-level tokens are making contribution, we also replace the scene and region tokens with the average pooling representation of that area. We observe that although having the same number of tokens, using average pooling, instead of scene region compression tokens will decrease both of the Lingo-Judage and the BLEU score, which indicates the effectiveness of the high-level tokens. Overall, these results confirm that applying multi-scale scene–region compression to early frames effectively balances efficiency and accuracy for autonomous driving video QA.

\begin{table}[]
\centering
\begin{tabular}{c|c|c}
\toprule
 \textbf{Model} & \textbf{ItoT} & \textbf{TtoI} \\
\midrule
CLIP ViT & 84.3 & 84.4 \\
SRC-ViT (CLIP-based) & 86.5 & 87.1  \\
\bottomrule
\end{tabular}
\caption{Stage one training evaluation.}
\label{Stage1_score}
\end{table}

\begin{table}[]
\centering
\begin{tabular}{c|c}
\toprule
 \textbf{Model} & \textbf{Lingo-J} \\
\midrule
QWen2VL & 51.15 \\
SRC-Pipeline & 52.40  \\
\bottomrule
\end{tabular}
\caption{Stage two zero-shot comparison.}
\label{zero-shot}
\end{table}

\begin{table}[h]
\centering
\small
\setlength{\tabcolsep}{6pt}
\renewcommand{\arraystretch}{1.2}
\begin{tabular}{c||c|c}
\toprule
\textbf{Model}   & \textbf{Ling-Judge}  & \textbf{BLEU} \\
\midrule
SRC-Pipeline & 57.28 & 12.09 \\
SRC-Pipeline (reverse)  & 41.02 & 8.64   \\
\bottomrule
\end{tabular}
\caption{Experiment on temporal importance.}
\label{temporal}
\end{table}

\subsection{Analysis Experiments}
In this section, we propose some analytic experiments on SRC-Pipeline, examine the effect of stage one training and our hypothesis that later frames' detail are more important than that of the early frames.
\subsubsection{Stage One Evaluation}
In stage one, we fine-tuned the SRC-ViT based on the QWen2VL's ViT backbone. To evaluate how the fine-tuning affect the performance of the ViT model, we utilize the ViT from CLIP-L/14 as the backbone to implement SRC-ViT, and evaluate its image-to-text (ItoT) retrieval and text-to-image (TtoI) retrieval performance with a corresponding CLIP-L/14's text encoder on a subset of CC3M dataset, which composes of 2000 data samples. The results are presented in Table \ref{Stage1_score}. From the table, we observe that the SRC-ViT outperform the vanilla CLIP's ViT on both of the ItoT and TtoI retrieval, indicating that the scene and region training procedure can boost the performance of vision encoder itself. This is also verified in Table \ref{zero-shot}, where we conduct a zero-shot evaluation of the original QWen2VL model and the SRC-Pipeline (The pre-trained QWen2VL model in which we replace the ViT to the fine-tuned SRC-ViT). We observe that there's a slightly improvement on the Lingo-J metric, which means the stage-one training will not harm, but increase the performance of the vision encoder.

\subsubsection{Temporal Hypothesis Verification}
The novelty of SRC-Pipeline is based on the hypothesis that in autonomous driving video question answering task, later frames' detail is more important than that of early frames. To verify the hypothesis, instead of using scene-region tokens for early four frames and leave the last one as patch tokens, we try a "reverse" implementation by using scene-region tokens for later four frames and leave the first one frame as patch tokens. By doing this, we remove the detailed semantic information of the latest frame, which should be the most important one in our hypothesis, which keep the total number of visual token unchanged. Base on the result on Table \ref{temporal}, we observe that there is a significant drop of performance between the original and "reverse" approach. Such a performance gap indicates that the detailed information of latest frames are important and cannot be removed. 

\section{Conclusion}
In this work, we introduced the SRC-Pipeline, an efficiency-driven framework for video question answering in autonomous driving. By compressing early frames into multi-scale scene and region tokens while preserving detailed patch representations for later frames, our approach significantly reduces FLOPs in the vision–language pipeline without sacrificing reasoning accuracy. Experiments demonstrate that both scene- and region-level tokens provide complementary benefits, enabling the model to retain global semantics and localized details while operating at one-third of the computational cost. This balance between efficiency and accuracy is particularly crucial for autonomous driving, where real-time response is essential for safety. While SRC-Pipeline achieves promising result on QWen2VL backbone, we plan to extend our scene\&region compression strategy to other backbones, using other autonomous driving training datasets, and perform evaluation with longer autonomous driving video streams in our future work.

{
    \small
    \bibliographystyle{ieeenat_fullname}
    \bibliography{main}
}

\end{document}